\documentclass[runningheads]{llncs}
\usepackage[T1]{fontenc}
\usepackage{graphicx}
\usepackage{booktabs}
\usepackage[misc]{ifsym}

\usepackage{mwe}

\usepackage{amssymb,amsmath,array}
\usepackage{tikz}
\usetikzlibrary{arrows,shapes}
\usetikzlibrary{patterns}
\usetikzlibrary{decorations.pathreplacing, patterns}
\usepackage{tabularx}
\usepackage{multicol}
\usepackage{multirow}
\usepackage{algpseudocode}
\usepackage{hyperref}
\usepackage{listings}
\usepackage[small]{caption}
\usepackage{algorithm}
\usepackage[switch]{lineno}
\usepackage{bbm}

\usepackage{multirow}%
\usepackage{xcolor}
\usepackage{multicol}

\newfloat{algorithm}{t}{lop}
\algnewcommand\algorithmichyperparams{\textbf{Hyperparameters:}}
\algnewcommand\AHyperparams{\item[\algorithmichyperparams]}

\algnewcommand\algorithmicparams{\textbf{Parameters:}}
\algnewcommand\AParams{\item[\algorithmicparams]}

\algnewcommand\algorithmicinput{\textbf{Input:}}
\algnewcommand\algorithmicoutput{\textbf{Output:}}
\algnewcommand\Input{\item[\algorithmicinput]}
\algnewcommand\Output{\item[\algorithmicoutput]}
\algnewcommand\algorithmicforeach{\textbf{for each}}
\algdef{S}[FOR]{ForEach}[1]{\algorithmicforeach\ #1\ \algorithmicdo}

\begin{document}

\title{Synthesizing real-world distributions from~high-dimensional Gaussian Noise with~Fully~Connected~Neural~Network}
\titlerunning{Synthesizing real-world distributions from Gaussian Noise}

\author{Joanna Komorniczak}

\authorrunning{J. Komorniczak}

\institute{\textit{Department of Systems and Computer Networks},\\ Wrocław University of Science and Technology}

\maketitle              

\begin{abstract}
The use of synthetic data in machine learning applications and research offers many benefits, including performance improvements through data augmentation and privacy preservation of original samples. This work proposes an efficient synthetic data generation method based on a fully connected neural network that transforms a high-dimensional random Gaussian distribution to approximate a target real-world dataset. The proposed solution combines data preprocessing designed for tabular data with distribution modeling and PCA dimensionality reduction to further enhance data privacy. The work also defines two dedicated randomized loss functions based on Wasserstein distance combined with feature Covariance and a randomized pairwise error reduction loss function. The experiments conducted on 25~diverse tabular real-world datasets confirm that the proposed solution obtains similarity and privacy scores relative to the state-of-the-art generative methods and achieves reference MMD scores orders of magnitude faster than modern deep learning solutions. The experiments involved analyzing distributional similarity, privacy protection, and the utility of synthetic data in classification tasks.

\end{abstract}

\section{Introduction}
The prevalence of data acquisition, transmission, and processing in the modern world has had a significant impact on the privacy and security of user data \cite{binjubeir2019comprehensive}. Data privacy has become a critical issue not only in medical applications \cite{guillaudeux2023patient} but also in many other domains, such as automatic transaction approval \cite{shao2020credit}, Internet of Things applications \cite{alabdulwahab2024privacy}, and even across social media platforms \cite{liu2021machine}.

Many of those sectors employ intelligent systems that involve machine learning models trained on original, private data. Recent years have highlighted the vulnerabilities of machine learning models, revealing that private user data used in the training process can be extracted from the final system \cite{chen2025survey}. Shokri et al. \cite{shokri2017membership} used Membership Inference Attacks to indicate whether a specific sample was involved in training a neural network model. The work on Inversion Attacks \cite{fredrikson2014privacy} has shown that patient genotype can be reconstructed from the trained model. Later, those attacks also demonstrated vulnerabilities in computer vision, enabling the reconstruction of face images from a trained face recognition model \cite{zhang2020secret}. Generative Regression Neural Network \cite{ren2022grnn} is another attack that can recover high-resolution images based on neural network gradients. Those works emphasize that machine learning models tend to memorize the training data~\cite{yeom2018privacy} and highlight the need for private data protection. New regulations further address those issues, requiring the anonymization of private data \cite{almeida2024umap}. While machine learning systems are deliberately intended to solve actual problems and process real-world datasets, the use of synthetically generated data has recently become a priority to prevent access to sensitive training data \cite{hu2022membership}.

In addition to the benefits associated with the use of synthetic data in privacy preservation, many data augmentation techniques allow for the improvement of classification quality in imbalanced domains \cite{dou2023machine}, the reduction of overfitting risk \cite{shorten2019survey}, or the extension of datasets for supervised learning tasks when the cost of human annotation or data acquisition is significant \cite{kim2021lada}.

\paragraph{Contribution}

This work proposes the \textit{Distribution Mapping with Shuffled Optimization} (\textsc{dimso}) -- a time-efficient synthetic data generation tool that transforms high dimensional Gaussian noise towards the target real-world distribution. Instead of encoding the original data into a latent space with autoencoders, the proposed \textsc{dimso} samples a random distribution $\mathcal{N}(0,1)$ and learns a mapping function that matches noise with a target distribution using a fully connected neural network architecture. The proposed method employs pre-processing and post-processing steps intended for categorical features and performs Principal Component Analysis (\textsc{pca}) to increase the privacy of original samples, and reduce memory and computational complexity.

The main contributions of the presented work include:
\begin{itemize}
    \item Proposition of a time-efficient method that employs a fully connected neural network to model real-world distributions.
    \item Definition of two loss functions that (a) combine per-feature Wasserstein distance with covariance differences and (b) employ a randomized pairwise sample distance.
    \item Comparison with traditional and deep-learning-based state-of-the-art generators regarding (a) their ability to model the target distribution, (b) privacy protection, and (c) the influence on recognition performance.
    \item Comparison of the execution time of the proposed methodology with state-of-the-art deep learning generators.
\end{itemize}

Compared to the neural-network-based generative models in the literature, the proposed approach focuses on simplicity and time efficiency. Compared to classic baselines, it incorporates randomness and \textsc{pca}-based projections into the generation process to improve privacy preservation.

\section{Related works}

Synthetic data generation offers many benefits in today's machine learning. The primary goals of synthetic data usage include augmentation to improve a model's generalization ability \cite{maharana2022review}, increasing the fairness of intelligent systems \cite{sharma2020data}, and preserving data privacy \cite{xu2019modeling}. Even though many modern generative solutions focus on generating unstructured data for computer vision or natural language processing \cite{shorten2021text}, there are many applications where generating tabular datasets is highly advantageous \cite{hernandez2022synthetic}.

\subsection{Data augmentation}
Modern machine learning, with recent advances in deep learning solutions, is strongly data-driven \cite{chakraborty2024machine}. Meanwhile, many domains to which such sophisticated methods are intended to be applied suffer from data scarcity \cite{safonova2023ten} or the high cost of sample acquisition, especially for rare events or minority classes \cite{dou2023machine}.

Augmentation techniques supplement the learning dataset with additional samples, usually aiming to resist model overfitting or aid learning without the additional cost of data collection \cite{mumuni2022data}. Augmentation techniques are frequently used in deep learning image classification. Basic image augmentation techniques used in computer vision include rotation, translation, filtering, and contrast changes, but deep learning approaches such as \textsc{gan}s and adversarial training are also used \cite{shorten2019survey}. 

In tabular data, augmentation is more challenging because the samples lack an invariant structure like images or text \cite{fang2022semi}. Tabular data often involves manual sample collection. The feature values may span a wide range, take various forms (e.g., categories), or exhibit complex dependencies \cite{wang2024challenges}. Existing tabular data augmentation solutions may implement various mechanisms to achieve the initial goal of the augmentation procedure. A recent review notes that data augmentation aimed at balancing an initially imbalanced classification problem may involve different manipulations than when the goal is to improve the fairness of the trained model or to increase the dataset size \cite{fonseca2023tabular}. Histogram Augmentation Technique (\textsc{hat}) \cite{sathianarayanan2022feature} is a general-purpose tabular data augmentation technique that aims to preserve the original distribution of samples. This method applies different generation mechanisms depending on whether a specific feature has a continuous or discrete distribution.

\subsection{Class imbalance}

Many approaches to data augmentation focus on imbalanced problems and perform oversampling to improve the representation of the minority class \cite{gosain2017handling}. Imbalanced datasets pose challenges to canonical machine learning methods \cite{lorena2019complex}. In cases of significant imbalance, the concept of an underrepresented minority class may be disregarded by the learning method. Given that the minority class is often more critical to recognize, e.g., in medical diagnostics or fraud detection \cite{shao2020credit}, many data augmentation solutions have been proposed to balance the dataset by synthesizing minority class examples.

Among the canonical oversampling methods, one should mention the \textit{Synthetic Minority Over-sampling Technique} (\textsc{smote}) \cite{chawla2002smote}, which generates synthetic data by linearly interpolating minority-class samples. Throughout the years, \textsc{smote} has become a standard approach to learning from imbalanced data, along with its extensions designed for specific purposes \cite{fernandez2018smote}. Even though the initial algorithm was proposed as a solution to improve classification quality due to class imbalance, some recent papers note that \textsc{smote} and its variants may outperform deep generative approaches in the synthetic tabular data generation task \cite{wang2024challenges}, demonstrating strong distributional similarity between real-world and synthetic samples.

However, \textsc{smote} has certain limitations, as linear interpolations may fail to capture multimodal distributions or complex feature dependencies present in the original data \cite{elreedy2019comprehensive}. Other research notes that the distribution of synthetic samples does not exactly follow the original minority class distribution, which may affect classification performance \cite{elreedy2024theoretical}.

\textsc{Smote}, although canonical, is not the only approach to imbalanced data oversampling. Generative Adversarial Networks (\textsc{gan}) \cite{mariani2018bagan} and Variational Autoencoders (\textsc{vae}) \cite{fajardo2021oversampling} are deep-learning-based solutions that have been shown to effectively balance classification problems and improve recognition performance. \textsc{Gan} comprises two neural network models -- a generator and a discriminator \cite{goodfellow2014generative}. The generator aims to produce synthetic samples that would be classified by the discriminator as real ones, drawn from the actual dataset. \textsc{Vae}s first encode the real-world data into a specific (usually Gaussian) distribution in a latent space and then generate data by sampling the embeddings and decoding them back to their original dimensionality \cite{kingma2013auto}. Despite the limitations of these generative models -- including the problem of mode collapse \cite{srivastava2017veegan} -- they are currently state-of-the-art in synthetic data generation. Later modifications of these methods, such as \textsc{sb-gan} and \textsc{sb-vae} \cite{akritidis2023conditional}, add additional steps to synthetic data generation. The authors of these modifications observed that the distribution of the minority class is significantly affected by outliers and noise, and they demonstrated that carefully selecting the instances used to train the models improves their performance. Combining \textsc{gan} architecture with the analysis of samples in the frequency domain was also an interesting oversampling solution to the problem of detecting fraud in credit card transactions \cite{shao2020credit}.

\subsection{Data privacy}

In addition to improving classification quality in the presence of data scarcity and class imbalance, synthetic data plays a significant role in the context of privacy preservation \cite{jordon2018pate}.

\textsc{smote} is a simple and effective baseline for data augmentation, however, it has significant limitations regarding data privacy. Recent research has shown \cite{ganev2025smote} that privacy attacks, such as the Membership Inference Attack, as well as Distin\textsc{smote} and Recon\textsc{smote}, designed specifically with the \textsc{smote} baseline in mind, allow for the reconstruction of the original training samples.

Due to the limitations of the baseline solution, \textsc{smote-dp} \cite{zhou2025smote} combines \textsc{smote} with a differentially private data generator. The presented study used the Bayesian network to satisfy differential privacy  \cite{zhang2017privbayes}. The authors of \textsc{umap-smotenc} \cite{almeida2024umap} combine \textsc{smote-nc} and \textsc{umap} projection to preserve the privacy of the data while efficiently synthesizing new samples. The solution compresses the samples into the two-dimensional space using Manifold Approximation and Projection \cite{mcinnes2018umap} prior to performing synthetic sampling. In contrast to the primary motivation for using \textsc{smote}, the datasets generated by \textsc{umap-smotenc} retain the original class proportions, hence, they may involve imbalance. 

As noted by the authors of \textsc{umap-smotenc}, the data sampling in reduced dimensionality can improve data privacy. A canonical dimensionality reduction method, Principal Component Analysis, can also transform the data into a lower-dimensional space to reduce the memory and computational cost of model training and improve classification quality \cite{reddy2020analysis}. While the primary components of original \textsc{pca} tend to provide much information about the original samples, some \textsc{pca} modifications focus directly on preserving data privacy \cite{froelicher2023scalable,fan2021ppca}.

Among the traditional machine learning models, one should also mention the Avatar method \cite{guillaudeux2023patient}, which generates a local model to produce synthetic data of mathematically simulated individuals. The method can use \textsc{pca} or other projection techniques to compute pairwise distances between samples and define the local area.

Similar to oversampling, many solutions for synthetic data generation with privacy preservation in mind rely on the use of \textsc{gan}s and \textsc{vae}s \cite{sharma2024generative}. The Table\textsc{gan} method \cite{park2018data} uses \textsc{gan}s combined with convolutional neural networks. \textsc{Tgan} additionally employs \textsc{lstm} with an attention mechanism \cite{xu2018synthesizing} to capture correlations between features. Med\textsc{gan} \cite{choi2017generating} combines an auto-encoder and \textsc{gan}s to simulate realistic data for medical purposes. The current state-of-the-art methods for tabular data are Conditional Tabular \textsc{gan} (\textsc{ctgan}) and Tabular \textsc{vae} (\textsc{tvae}) \cite{xu2019modeling}, which, due to their efficient, open-source implementations \cite{patki2016synthetic}, still serve as a baseline for data generation and comparison with reference approaches. The authors of these solutions challenge traditional generative models to capture the specifics of tabular data, such as mixed data types or non-Gaussian, multimodal distributions. Both \textsc{ctgan} and \textsc{tvae} use fully connected neural network structures.

A solution directly focused on differential privacy is \textsc{pate-gan} \cite{jordon2018pate}, which replaces the discriminator component of the \textsc{gan} with Private Aggregation of Teacher Ensembles, resulting in asymmetrical training involving the generator, discriminator-teachers, and discriminator-students. \textsc{Dp-gan} \cite{ho2021dp} later introduced a third component of \textsc{gan}, denoted as a differential privacy identifier, to ensure data privacy. Other solutions with a similar motivation combine \textsc{ctgan} with controlled noise injection \cite{alabdulwahab2024privacy}

Solutions based on \textsc{vae}s have also been proposed. The Oblivious Variational Autoencoder \cite{vardhan2020synthetic} uses differentiable oblivious decision trees in both the encoder and decoder structures of the \textsc{vae}. The decision trees are aggregated into layers to capture complex data distributions. Generative\textsc{mtd} \cite{sivakumar2023generativemtd}, designed for small datasets, uses a \textsc{vae-gan} architecture trained on pseudo-real samples generated based on the original sample's nearest neighbors. Tab-\textsc{vae} \cite{tazwar2024tab} distinguishes between types of features before encoding the data and employs a variational Gaussian mixture model to capture multimodal distributions. \textsc{Ttvae} \cite{wang2025ttvae} employs a transformer architecture instead of fully connected layers to capture complex feature dependencies. 

One of the drawbacks of those deep learning solutions is the memory complexity and computational cost involved in their training and employment \cite{miletic2025utility}. Moreover, as is typical for deep learning solutions, they often require large training datasets \cite{wang2024challenges}. A good alternative is a more traditional statistical solution, e.g., Gaussian Copula \cite{patki2016synthetic}, which models each continuous feature as a Gaussian distribution and each categorical feature as a Truncated Gaussian distribution. Synthpop \cite{nowok2016synthpop} models conditional distributions of features using \textsc{cart} decision trees.

\section{Distribution Mapping with Shuffled Optimization}

\begin{figure}[!htb]
    \centering
    \includegraphics[width=\linewidth]{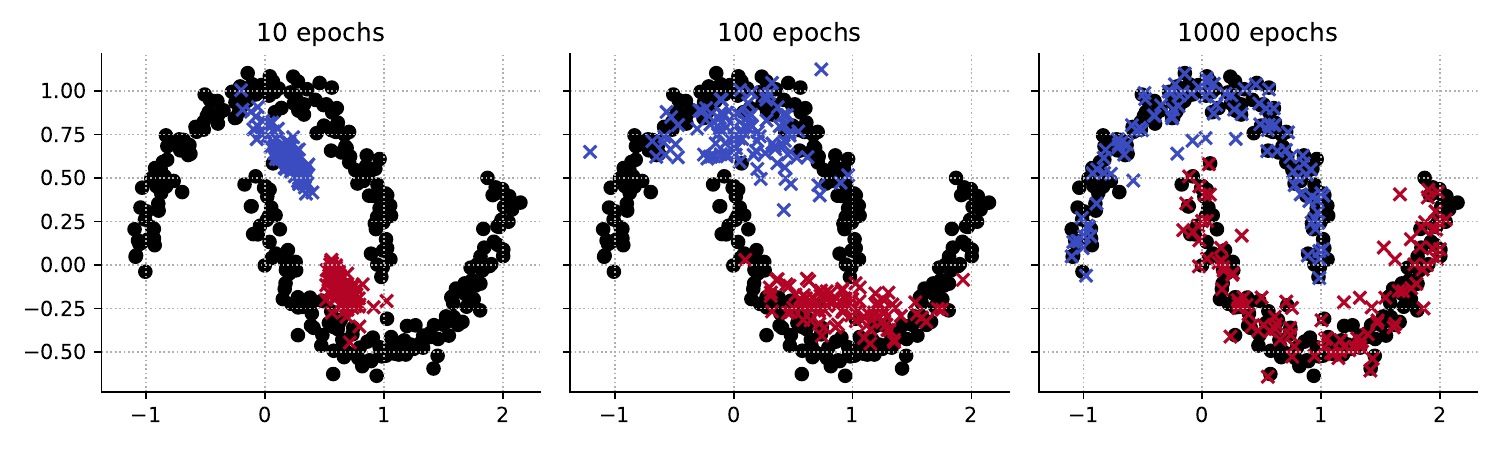}
    \caption{The iterative process of distribution matching towards a \textit{moons} dataset (black points). The red and blue markers represent two classes from synthetic data after 10, 100, and 1000 epochs.}
    \label{fig:moons-rae}
\end{figure}

This work presents a \textit{Distribution Mapping with Shuffled Optimization} (\textsc{dimso}) generation tool for synthetic data generation that transforms random high-dimensional Gaussian noise into a target real-world distribution using a fully connected neural network. The code of the method is publicly available\footnote{\url{https://github.com/JKomorniczak/DiMSO/}}. The optimization process uses randomized loss functions, where subsets of samples are selected at each optimization step, and \textsc{pca} feature extraction to reduce memory and time complexity while preserving data privacy. Figure \ref{fig:moons-rae} shows the synthetic distribution after 10, 100, and 1000 epochs of model training. The initial epochs show vaguely shifted normal distributions that, with subsequent epochs, start to align with the target. At the final step of 1000 epochs, the synthetic distribution is almost perfectly aligned with the target data.

\subsection{Procedure}

The initial random distribution has a higher dimensionality than the target. While this dimensionality increases the number of neural network parameters, impacting computational complexity, the preliminary experiments have shown that providing such flexibility benefits distribution modeling capabilities. Distribution transformation is performed for each problem class separately. The number of samples drawn from the normal distribution prior to mapping is specified by a parameter and determines the number of samples in a given class after the optimization procedure. By default, the initial class proportions are maintained in the problem. Pseudocode \ref{alg:alg} describes the operation of the proposed approach.

\begin{algorithm}[h]
\scriptsize
\caption{\textit{Distribution Mapping with Shuffled Optimization}}
\label{alg:alg}
\begin{algorithmic}[1]

\Input
\Statex labeled dataset $(X, y)$

\AHyperparams
\Statex
$f$ -- features factor,
$n_s$ -- number of synthetic samples in a given class,
$b$ -- preserve class imbalance boolean flag,
$p$ -- perform PCA boolean flag,
$v$ -- retained PCA variance,
$\mathcal{L}$ -- loss function,
$e$ -- number of epochs

\AParams
\Statex
$\mathcal{S}$ -- per-class source distributions,
$\mathcal{M}$ -- per-class neural networks,
$\mathcal{T}$ -- per-class target distributions

\Output
\Statex
$(X_{syn}, y_{syn})$ -- generated dataset

{\color{gray}\Comment{\textsc{preprocess}}}
\State $X \gets \textsc{Preprocess discrete features}(X)$
\State $X \gets \textsc{Normalize features}(X)$

\State $X \gets \textsc{PCA}(X)$ retaining variance $v$ according to $p$
\State $\mathcal{T} \gets $ split $X$ into class subsets $\mathcal{T}_i$

{\color{gray}\Comment{\textsc{initialize source distributions}}}
\State noise dimensionality $d \gets \lfloor f \cdot \mathrm{dim}(X)\rfloor$
\State $(n_0, n_1, ... n_k) \gets$ source size for each class according to $b$ and $n_s$

\For{each class $i$ in $0,1 \ldots k$}
    \State sample $\mathcal{S}_i \sim \mathcal{N}(0,1)$ of size $(n_i,d)$
    \State initialize neural network $\mathcal{M}_i:d\rightarrow \mathrm{dim}(X)$
    \EndFor

{\color{gray}\Comment{\textsc{training}}}
\For{each class $i$ in $0,1 \ldots k$}
    \State $\mathcal{M}_i \gets$ \textsc{Optimize} $\mathcal{M}_i(\mathcal{S}_i)$ with $ \mathcal{L}(\mathcal{S}_i,\mathcal{T}_i)$ for $e$ epochs
\EndFor

{\color{gray}\Comment{\textsc{generate synthetic data}}}
\State $X_{syn}\gets\emptyset$
\State $y_{syn}\gets\emptyset$
\For{each class $i$ in $0,1 \ldots k$}
    \State optionally sample new Gaussian noise $\mathcal{S}_i$
    \State $X_i \gets \mathcal{M}_i(\mathcal{S}_i)$
    \State $X_{syn}\gets X_{syn}\cup X_i$
    \State $y_{syn}\gets y_{syn}\cup y_i$
\EndFor

{\color{gray}\Comment{\textsc{post-process}}}
\State $X_{syn} \gets \textsc{Inverse PCA}(X_{syn})$ according to $p$
\State $X_{syn} \gets \textsc{Revert normalization}(X_{syn})$
\State $X_{syn} \gets \textsc{Restore discrete features}(X_{syn})$
\State randomly shuffle samples in $(X_{syn},y_{syn})$
\State \Return $(X_{syn},y_{syn})$

\end{algorithmic}
\end{algorithm}

The algorithm takes the real-world classification dataset ${X, y}$ as input. The hyperparameters of \textsc{dimso} include the features factor $f$ which impacts the Gaussian noise dimensionality, $n_s$, indicating the number of samples per class, and two boolean flags indicating if \textsc{pca} is performed ($p$) and if the class proportions are maintained ($b$). The remaining hyperparameters describe neural network training: the number of epochs $e$, the learning rate $\eta$, and the loss function $\mathcal{L}$.

The first steps involve data preprocessing, including the detection of categorical features and the values they take, as well as standard normalization of each feature. If the hyperparameter $p$ is set to \textit{True}, the \textsc{pca} dimensionality reduction is performed to the number of features that explain at least $v$ of data variance. The target distribution is also divided into specific classes and stored in $\mathcal{T}$. These steps are described in lines 1:4 of the pseudocode.

Initial dimensionality $d$ of random normal distributions $\mathcal{S}$ is set according to the $f$ hyperparameter and the dimensionality of a target $X$. The method initializes a set of neural network models, $\mathcal{M}$, containing a separate model for each problem class in $y$. The neural network has $d$ inputs, and the number of outputs is set according to the dimensionality of $X$. This part of the procedure is described by lines 5:10. The network optimization is performed for each class separately, aiming to map the initially random distribution $\mathcal{S}_i$ towards the $i$-th class distribution $\mathcal{T}_i$ with a model $\mathcal{M}_i$. The criterion is defined by the loss function $\mathcal{L}$. The model weights are updated for a specified number of epochs $e$, as described in lines 11:13 of the pseudocode.

After the optimization, the per-class distributions transformed with each model in $\mathcal{M}$ are combined to form a final synthetic distribution $X_{syn}$. The label vector $y_{syn}$ is also incrementally created, as described in lines 14:21. The final steps include the post-processing (lines 22:24) with the inverse of \textsc{pca}, rescaling the data to the original range, and discretizing categorical features. The algorithm returns a multiclass-synthetic dataset with a random order of samples.

\subsection{Components of the method}

The proposed approach employs a fully connected neural network with three hidden layers of 100 neurons each and a ReLU activation function. In optimization, the Adam optimizer was used, and a constant learning rate of $\eta = 10^{-3}$. 

The method transforms an initially random distribution towards a target one by optimizing a specific loss function, which is a primary factor that impacts the optimization process. The default loss function is based on the absolute error between randomly assigned pairs of samples from the synthetic distribution \textit{S} and the target distribution \textit{T}. Such a loss function $\mathcal{L}_{RAE}$ is described in Equation \ref{eq:rae-loss}, with $n$ indicating the number of samples and $d$ -- the data dimensionality.

\begin{equation}
\mathcal{L}_{RAE}(S, T) 
= \sum_{i=1}^{n} \sum_{j=1}^{d} 
\left| S_{\pi(i)}^{j} - T_{\pi(i)}^{j} \right|
\label{eq:rae-loss}
\end{equation}

$\mathcal{L}_{RAE}$ aims to directly match the pairs of samples from synthetic and target distributions, the reference loss function uses Wasserstein distance and feature Covariance, as described in Equation \ref{eq:wc-loss}.

\begin{equation}
\mathcal{L}_{WC}(S, T)
= \frac{WD(S, T) + Cov(S, T)}{2}
\label{eq:wc-loss}
\end{equation}

The raw $WD(S, T)$, described in Equation \ref{eq:wd} computes the distance in each dimension separately, hence, it does not consider the relationship between problem attributes. Using the feature Covariance $Cov(S, T)$, described in Equation \ref{eq:cov}, as an additional optimization factor should enhance a method's ability to adequately represent the actual real-world problem.

\begin{equation}
WD(S, T) 
= \sum_{j=1}^{d} \frac{1}{N} \sum_{i=1}^{N} 
\left| S_{\pi(i)}^{j} - T_{\pi(i)}^{j} \right|
\label{eq:wd}
\end{equation}

\begin{equation}
Cov(S, T)
= \sum_{i=1}^{d} \sum_{j=1}^{d} 
\left| \mathrm{Cov}(S)_{\pi(i)}^{j} - \mathrm{Cov}(T)_{\pi(i)}^{j} \right|
\label{eq:cov}
\end{equation}

Both loss functions use a random subset of data instances, $\pi$. This is a characteristic component of the proposed \textsc{dimso} approach. While the order of samples in $\pi$ does not impact the values of the Wasserstein distance and Feature Covariance, it will play a significant role in \textsc{rae} loss. Since the assignment of pairs is random, the optimization will face many conflicting criteria while transforming the multidimensional hyperplane of the synthetic random distribution. Given the low computational complexity of the loss itself and a simple neural network architecture, the cost of a non-optimal order of $\pi$ will not affect the optimization process and will offer the benefit of non-ideal sample overlap, which is crucial for privacy preservation in synthetic data. The number of samples in the $\pi$ subset is set according to the number of samples in a given class in the real-world dataset and a synthetic one, selecting the minimum of those two values.

\begin{figure}[!htb]
    \centering
    \includegraphics[width=\linewidth]{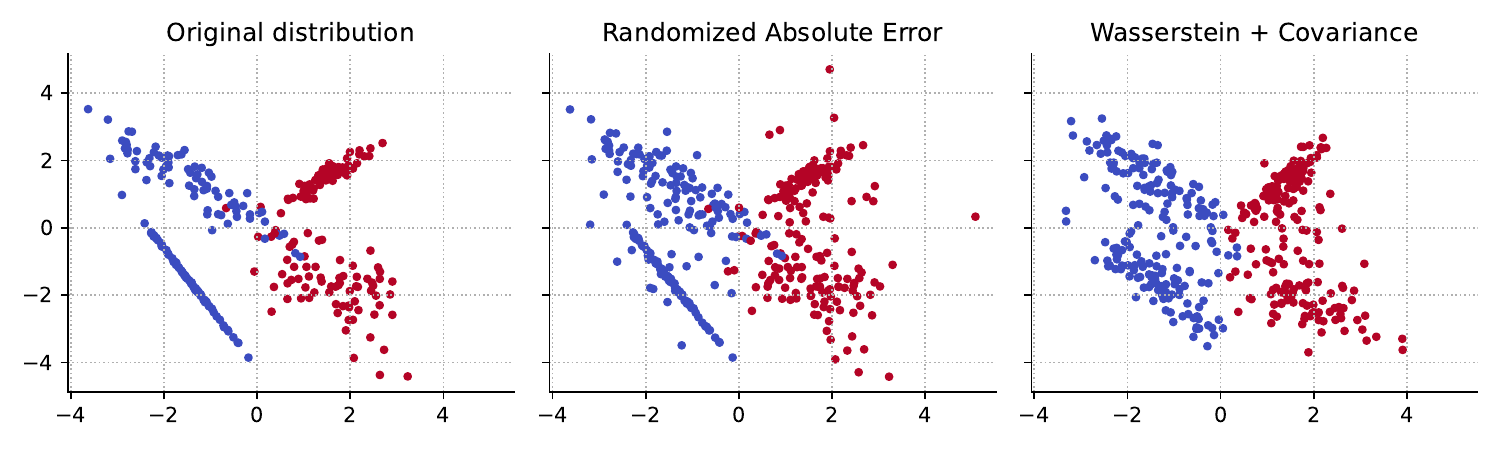}
    \caption{The illustrative example of synthetic data distribution matching towards a target distribution (left) after 500 epochs and the use of different loss functions -- $\mathcal{L}_{RAE}$ (center) and $\mathcal{L}_{WC}$ (right).}
    \label{fig:make-clf}
\end{figure}

The examples of distribution matching for a simple planar classification case are shown in Figure \ref{fig:make-clf}.

The Figure shows the results of optimization after 500 epochs for $\mathcal{L}_{RAE}$ (center), and a loss function that combines the Wasserstein Distance $WD$ and feature covariance $Cov$ in a $\mathcal{L}_{WC}$ (right).

\subsection{Optimizing to principal component's distribution}

To enhance privacy preservation and reduce memory complexity, the proposed method uses \textsc{pca} dimensionality reduction and maps the initial random noise to the principal components of the real-world dataset. The hyperparameter $v$ describes the percentage of variance that is aimed to be explained by \textsc{pca}. Small values will increase privacy but may impact the similarity score of the generated data. Values close to 100\% will result in synthetic data that reliably represent the real-world distributions but may reveal sensitive data. After optimization, the features are inverse transformed to the original dimensionality. 

\textsc{Pca} is widely used to improve classification quality \cite{reddy2020analysis} and reduce the computational complexity of learning systems \cite{velliangiri2019review}. After dimensionality reduction, the extracted features that lack semantic meaning are employed to train learning models. Using \textsc{pca} components instead of the original features can be viewed as a mechanism for preserving privacy \cite{banu2009preservation}.

\begin{figure}[!htb]
    \centering
    \includegraphics[width=\linewidth]{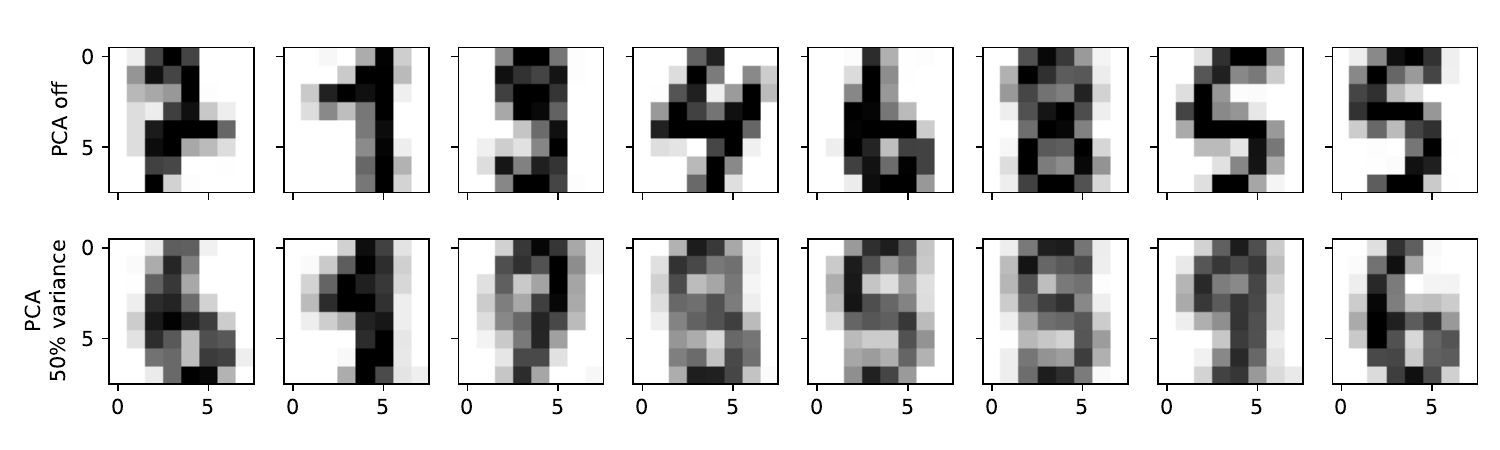}
    \caption{Visualization of samples generated in the original feature space (top row) and after inverse transform from \textsc{pca} components after generation in reduced dimensionality (bottom row).}
    \label{fig:rae-pca}
\end{figure}

Figure \ref{fig:rae-pca} visualizes the synthetic samples generated with \textsc{dimso} when optimizing to the original feature space (top row) and after \textsc{pca} dimensionality reduction (bottom row). In the second scenario, the synthetic samples were intended to imitate the extracted \textsc{pca} components that explained over 50\% of the variance. After the generation procedure, the synthetic components were subjected to the inverse transform to their initial dimensionality. In this example, the default \textsc{rae} loss function was used.

Even though the method was designed for tabular data, this example used a dataset that presents images of size 8x8. It was motivated by the possibility of visualizing the high-dimensional data (64 features) as images. The pixel intensities were flattened into a vector and treated as numeric tabular features. As shown in the Figure, the synthetic principal components after inverse transformation appear more blurry. This specific \textsc{pca} usage, in addition to reducing computational costs, can further enhance original data protection.

\section{Experiment design}

This section will describe the experimental environment, present the datasets, the protocol, and the goals of the experiments. The proposed approach and the experiments were implemented in Python programming language with \textit{scikit-learn} \cite{scikitlearn} and \textit{pytorch} libraries. The datasets were obtained with the \textit{pmlb} library \cite{romano2021pmlb}. Reference methods came from \textit{synthyverse} \cite{synthyverse2026} library.

\subsection{Datasets}\label{sub-datasets}

\begin{table}[!htb]
    \caption{Characteristics of real-world datasets used in experiments. The datasets were sorted from the most imbalanced (top rows) to balanced (bottom rows).}
    \vspace{0.5em}
    \centering
    \setlength{\tabcolsep}{10pt}
\renewcommand{\arraystretch}{1.0}
\scriptsize
\begin{tabular}{l|rrr|rr}
\toprule
 \textsc{dataset}            &   \# \textsc{samples} &   \# \textsc{features} &   \# \textsc{classes} &   \textsc{prior}$_{max}$ &   \textsc{prior}$_{min}$ \\
\midrule
allrep                &      3772 &         29 &         4 &    0.967 &    0.009 \\
 yeast                 &      1479 &          8 &         9 &    0.313 &    0.014 \\
 hypothyroid           &      3163 &         25 &         2 &    0.952 &    0.048 \\
lawsuit   &       264 &          4 &         2 &    0.928 &    0.072 \\
 backache              &       180 &         32 &         2 &    0.861 &    0.139 \\
 balance\_scale         &       625 &          4 &         3 &    0.461 &    0.078 \\
 cars                  &       392 &          8 &         3 &    0.625 &    0.173 \\
 mofn\_3\_7\_10           &      1324 &         10 &         2 &    0.779 &    0.221 \\
 schizo                &       340 &         14 &         3 &    0.588 &    0.182 \\
 breast\_cancer         &       286 &          9 &         2 &    0.703 &    0.297 \\
 penguins              &       333 &          7 &         3 &    0.438 &    0.204 \\
 xd6                   &       973 &          9 &         2 &    0.669 &    0.331 \\
 profb                 &       672 &          9 &         2 &    0.667 &    0.333 \\
 monk2                 &       601 &          6 &         2 &    0.657 &    0.343 \\
 saheart               &       462 &          9 &         2 &    0.654 &    0.346 \\
 tic\_tac\_toe           &       958 &          9 &         2 &    0.653 &    0.347 \\
 biomed                &       209 &          8 &         2 &    0.641 &    0.359 \\
 ionosphere            &       351 &         34 &         2 &    0.641 &    0.359 \\
 tokyo1                &       959 &         44 &         2 &    0.639 &    0.361 \\
 irish                 &       500 &          5 &         2 &    0.556 &    0.444 \\
 glass2                &       163 &          9 &         2 &    0.534 &    0.466 \\
 bupa                  &       345 &          5 &         2 &    0.510  &    0.490  \\
 parity5+5             &      1124 &         10 &         2 &    0.504 &    0.496 \\
germangss &       400 &          5 &         4 &    0.250  &    0.250  \\
 prnn\_crabs            &       200 &          7 &         2 &    0.500   &    0.500   \\
\bottomrule
\end{tabular}
    \label{tab:datasets}
\end{table}

The research used 25 publicly available real-world tabular datasets with a diverse number of features, samples, and imbalance ratios. A detailed description of the datasets is presented in Table \ref{tab:datasets}. While most of the datasets describe binary classification tasks, up to 9 classes were considered in a few multiclass problems. The last two columns of the table present the maximum and minimum prior probabilities of the classes, which affect the imbalance ratio of the problem. The datasets are sorted according to the proportion between minimum and maximum prior probabilities, presenting the most imbalanced datasets in the top rows. The datasets were characterized by diverse dimensionalities (from 4 to 44 features) and various numbers of samples (from small datasets with 163 to large ones, with over 3700 instances). 

In the experiments, we adopted the automatic detection of categorical features according to the number of values the specific features took. To allow the processing of digital datasets, where the values are saved on up to 4 bits, the number of values to consider a variable as categorical was set to 16. Such discrete feature detection approach is implemented as a component of \textsc{dimso} and was used for all other reference approaches.

\subsection{Metrics}\label{sub-metrics}

The experiments aimed to evaluate three primary objectives of the proposed \textsc{dimso} generator:
\begin{itemize}
    \item \textbf{the ability of the method to model the real-world distribution}, which was evaluated using the following metrics:
        \begin{itemize}
            \item The statistic of the Kolmogorov-Smirnov test (\textsc{ks}) comparing the distributions of the original and synthetic datasets computed independently per feature and averaged. Small values indicate similar distributions.
            \item Kullback–Leibler divergence (\textsc{kld}) quantifying the relative entropy between real and synthetic probability distributions, computed independently per feature and averaged. Small values indicate better similarity.
            \item Normalized mutual information (\textsc{nmi}) computed for every feature pair from the real and synthetic data and averaged. Higher scores indicate better preservation of feature dependencies.
            \item Maximum Mean Discrepancy (\textsc{mmd}) \cite{gretton2006kernel}, defined using the \textsc{rbf} kernel with $\gamma$ calculated dynamically according to the compared distributions.
        \end{itemize}
    \item \textbf{the privacy preservation}, which was evaluated using the following metrics:
        \begin{itemize}
            \item \textsc{Auc} of the membership inference attack (\textsc{mia auc}) using a Random Forest Classifier to distinguish between the synthetic and reference distributions. An \textsc{auc} of 50\% indicates high privacy, while higher values indicate more effective attacks \cite{khalil2025membership}.
            \item  Distance to Closest Record (\textsc{dcr}) considers the distances from synthetic samples to original samples. High values indicate better privacy.
            \item \textsc{Auc} of the Data Plagiarism Index membership inference attack (\textsc{dpi auc}) \cite{ward2024data}, which conducts an attack based on \textsc{dpi} values representing the synthetic and real instances in the neighborhood of samples. Similar to \textsc{mia auc}, results close to 50\% indicate high privacy. 
        \end{itemize}
    \item \textbf{the utility in the classification task} was evaluated with:
        \begin{itemize}
            \item Machine Learning Efficacy, defined as the difference between the \textsc{auc} of an \textit{xgboost} classifier trained with synthetic data and the \textsc{auc} of a model trained with original samples (\textsc{mle auc diff}). Values close to zero indicate high utility due to similar \textsc{auc} scores, and values above zero indicate the improvement of classification performance when synthetic data were used.
        \end{itemize}
\end{itemize}

The first set of measures quantifies how well the generated data reflects the real-world distribution. In evaluating distribution similarity, the ideal replication of the original data samples would yield the highest similarity score, indicating a potentially valuable method. However, such sample replication would not offer one of the most important benefits of synthetic data use, which is privacy \cite{theis2015note}. The second objective, data privacy preservation, therefore, conflicts with the first one. The last objective focuses on the impact on classification quality when synthetic data is used to train the classification model. This evaluation aspect is affected by the generator's ability to realistically approximate the real-world data \cite{jordon2018pate}.

\subsection{Reference approaches}\label{sub-references}

The experiments compared \textsc{dimso} to six reference generative methods. The complete description of the evaluated methods is presented in Table \ref{tab:hyperparam}. The references include three deep learning approaches: \textsc{ctgan} (representative of \textsc{gan} for tabular data), \textsc{tvae} (representative of \textsc{vae} for tabular data), and TabDiff (representative of the diffusion model for tabular data). Three classic approaches include \textsc{smote}, Univariate baseline, and Synthpop. The experiment compared those references with \textsc{dimso} in two configurations: with $\mathcal{L}_{RAE}$ and \textsc{pca} reduction, and with $\mathcal{L}_{WC}$ without \textsc{pca}. \textsc{Pca} was used in the case of \textsc{rae} to increase privacy preservation.

\begin{table}[!htb]
    \centering
    \caption{Description of evaluated synthetic data generation methods used in experiments and their hyperparameters.}
    \vspace{0.5em}
    \setlength{\tabcolsep}{5pt}
    \renewcommand{\arraystretch}{1.2}
    \scriptsize
    \begin{tabular}{p{2cm}|p{3.5cm}p{5.3cm}}
    \toprule
        \textsc{acronym} & \textsc{description} & \textsc{hyperparameters} \\ \midrule
        CTGAN & Conditional Tabular GAN \cite{xu2019modeling}& embedding\_dim=128, $\eta=0.0002$, $e=500$ epochs \\
        TVAE &  Tabular Variational Autoencoder \cite{xu2019modeling}&  embedding\_dim=128, $e=500$ epochs \\
        TabDiff & Mixed-type Diffusion Model for Tabular Data \cite{shi2025tabdiff} & $e=500$ epochs, $\eta=0.001$ \\ \midrule
        SMOTE & SMOTE / SMOTE-NC \cite{chawla2002smote} &  Default number of neighbors $k=5$, SMOTE-NC was used in case of categorical attributes\\ 
        Univariate & Univariate baseline generator & Each feature modeled independently\\
        Synthpop & Synthpop \cite{nowok2016synthpop} & Using sequential decision trees with minibucket=10, minibucket\_sampling=3, $K=100$\\\midrule
        DiMSO-R PCA & DiMSO with $\mathcal{L}_{RAE}$ and PCA reduction & $f=3.5$, $e=500$ epochs, $\eta=0.001$, $p$=True, $v=0.85$\\
        DiMSO-W & DiMSO with $\mathcal{L}_{WC}$ without PCA reduction &  $f=3.5$, $e=500$ epochs, $\eta=0.001$, $p$=False
    \\ \bottomrule
    \end{tabular}
    \label{tab:hyperparam}
\end{table}

\subsection{Goals of experiments}

The experiments focused on evaluating (a) the ability of \textsc{dimso} to simulate the real-world distribution, and (b) the method's processing time compared to deep learning references.

\paragraph{The ability to model real-world distribution while preserving privacy}

The first experiment directly evaluated the ability of \textsc{dimso} and the reference approaches described in Subsection \ref{sub-references} to model real-world distributions. The experiment used 25 diverse tabular datasets described in Subsection \ref{sub-datasets} and the metrics described in Subsection \ref{sub-metrics}. The experiment employed 5-fold stratified cross-validation. The generative model was fitted based on the training set, and the testing subset was used in the evaluation to calculate privacy and utility metrics. In this experiment, all methods maintained the class proportions that were present in the original dataset.

The available research has shown that classic methods, like \textsc{smote}, often produce better similarity results compared to sophisticated deep learning methods \cite{wang2024challenges}. However, the use of those simple baselines shows the risk of training data leakage. In this experiment, we expect to see a similar tendency. \textsc{Dimso}, which incorporates several randomization mechanisms, should demonstrate a better ability to preserve privacy than simple methods. The proposed loss functions used in \textsc{dimso} aim to produce synthetic data that reliably approximates the original distribution.

\paragraph{The processing time comparison}

The second experiment focused on the time complexity of the method. Since many modern generative models employ deep learning -- such as \textsc{vae}s or \textsc{gan}s -- their use often places significant computational demands. The proposed \textsc{dimso} uses a simple architecture and a loss function of low computational complexity. Compared with canonical deep learning models, the generation time with \textsc{dimso} is expected to be lower than that of reference complex architectures. In this experiment, we compared the generation times with \textsc{tvae}, \textsc{ctgan}, and TabDiff. The reference methods, configured as described in Table \ref{tab:hyperparam}, were used to generate data based on \textit{ionosphere} dataset, which contains 351 samples in 34 dimensions. The dataset describes a binary classification problem with 64\% samples representing the majority class. The generation procedure was replicated 5 times in a cross-validation procedure.

After data generation with reference deep learning solutions for a specified number of epochs ($e=500$), the \textsc{mmd} of the generated distribution was measured. \textsc{Dimso}, with two configurations specified in \ref{tab:hyperparam}, aimed to achieve the \textsc{mmd} of the deep learning solution in up to 1000 epochs. Once the target similarity metric was achieved, the optimization with \textsc{dimso} was stopped, and the processing time was saved for comparison. The experiments were conducted using a 14-core \textsc{cpu} with an \textsc{arm} architecture and a maximum clock rate of 4.51 GHz.

Since all methods evaluated in this experiment optimize the distributions in iterative mode (epochs), setting a target distribution similarity with \textsc{mmd} will reliably assess the time of the methods' operation and discard the factors related to their capability to model real-world data. Since \textsc{dimso} is a simpler model in terms of architecture, some datasets may require more epochs to achieve target similarity, however, the time per iteration should be lower compared to reference deep learning approaches.

\section{Experiment results}

This section presents and analyzes the results of the conducted experiments.

\subsection{The ability to model real-world distribution while preserving privacy}

\begin{figure}[!htb]
    \centering
    \includegraphics[width=\linewidth]{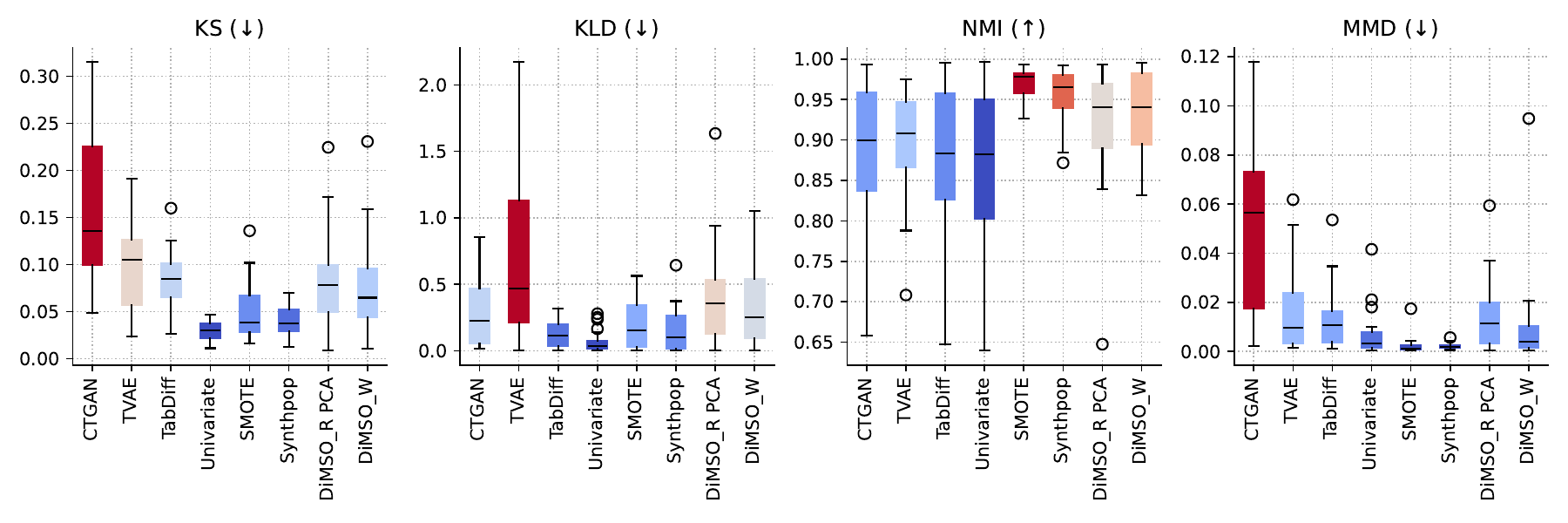}
    \caption{Similarity metric results for all evaluated datasets and compared methods. The bars represent the range of obtained scores for specific methods (x-axis) and their colors are dependent on the mean result across datasets. The lower means are indicated with blue and higher values with red. The direction in which we aim to optimize given metric is indicated in the description of a subplot.}
    \label{fig:e2-sim}
\end{figure}

\begin{figure}[!htb]
    \centering
    \includegraphics[width=\linewidth]{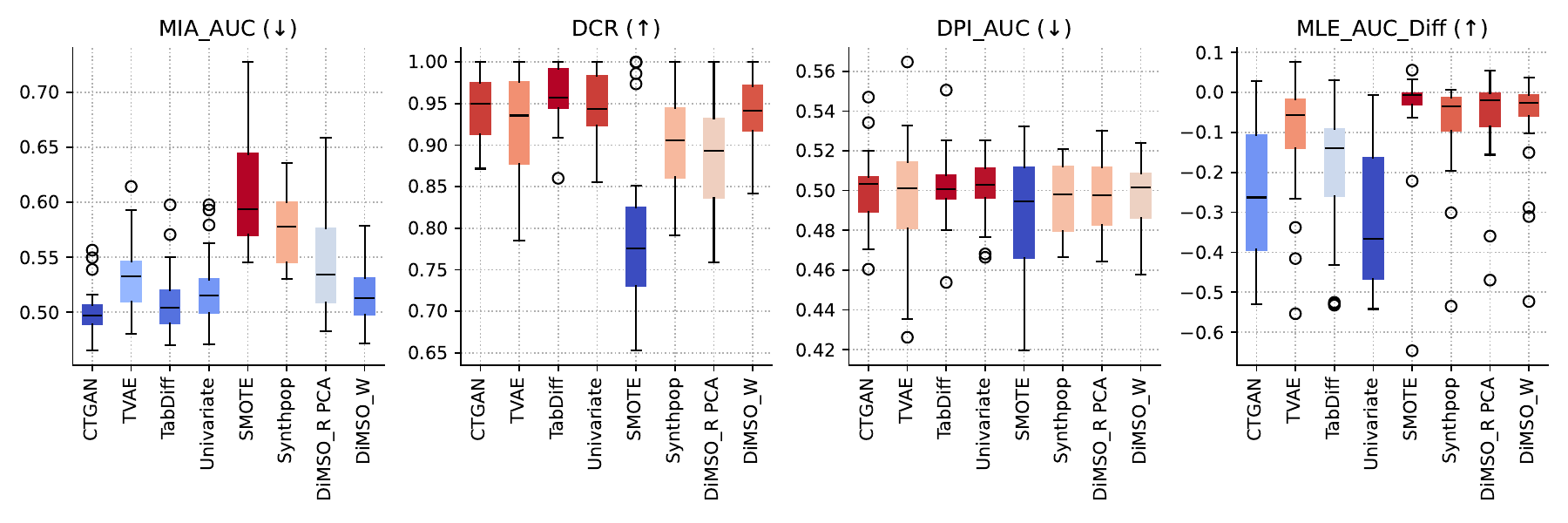}
    \caption{Privacy and utility metric results for all evaluated datasets and compared methods. The bars represent the range of obtained scores for specific methods (x-axis).}
    \label{fig:e2-priv}
\end{figure}

The results of the first experiment are presented in Figures \ref{fig:e2-sim} and \ref{fig:e2-priv}, with the first one presenting four metrics that describe distribution similarity and the second one presenting three metrics that describe privacy preservation and the utility metric. The figures present the range of values obtained across all datasets by the compared methods. The colors are used to additionally indicate the mean result for a specific method. Red indicates a high metric value, and blue indicates a low value.

As expected, the classic baselines (\textsc{smote}, Univariate, and Synthpop) achieve the lowest scores in \textsc{ks} and \textsc{kld} metrics. In the \textsc{ks}, \textsc{dimso} achieves a similar score to TabDiff and slightly better mean results than \textsc{ctgan} and \textsc{tvae}. In the \textsc{kld} metric, \textsc{ctgan} scores are relatively low, meanwhile, \textsc{tvae} shows low distribution similarity. Those two metrics are calculated for each feature independently, therefore, they do not consider the relation across features in a dataset. For that reason, the simple Univariate baseline shows very good results. \textsc{Nmi} reveals limitations of this baseline, making it the worst performing method. In this metric, the best results are still achieved by \textsc{smote} and Synthpop. The proposed \textsc{dimso} surpasses the deep-learning references, showing its ability to maintain relations across feature pairs. In the \textsc{mmd} metric, all methods except \textsc{ctgan} achieve good results. Simple baselines generate data with the best average similarity expressed in \textsc{mmd}. The precise results of this metric for specific datasets are presented in Table \ref{tab:mmd}. The cells represent scores averaged from cross validation folds, and the values in parentheses indicate the rank of the result. The table shows that \textsc{smote} and Synthpop produce the best \textsc{mmd} scores in most cases, however, for a few datasets, \textsc{dimso} and Univariate generators surpassed them.

\begin{table}[!htb]
    \centering
    \caption{The averaged results of \textsc{mmd} metric describing the distribution similarity. Lower scores indicate better result. The values in parenthesis indicate rank and the best result per dataset is emphasized with bold.}
    \vspace{0.5em}
    \setlength{\tabcolsep}{3pt}
\renewcommand{\arraystretch}{1.0}
\tiny
\begin{tabular}{l|rrrrrrrr}
\toprule
 dataset & CTGAN     & TVAE      & TabDiff   & Univariate & SMOTE     & Synthpop  & DiMSO-R & DiMSO-W   \\ \midrule
 allrep          & 0.013 (6) & 0.002 (3) & 0.027 (8) & 0.008 (5)  & \textbf{0.001} (1) & 0.001 (2) & 0.024 (7)   & 0.005 (4) \\
 yeast           & 0.070 (8) & 0.010 (5) & 0.010 (6) & 0.003 (4)  & 0.001 (3) & \textbf{0.001 }(1) & 0.014 (7)   & 0.001 (2) \\
 hypothyroid     & 0.012 (5) & 0.003 (3) & 0.022 (7) & 0.009 (4)  & \textbf{0.000} (1) & 0.001 (2) & 0.025 (8)   & 0.017 (6) \\
 lawsuit         & 0.118 (8) & 0.014 (6) & 0.011 (5) & 0.005 (4)  & \textbf{0.001} (1) & 0.004 (3) & 0.015 (7)   & 0.002 (2) \\
 backache        & 0.070 (7) & 0.021 (5) & 0.004 (2) & 0.004 (3)  & \textbf{0.001} (1) & 0.004 (4) & 0.037 (6)   & 0.095 (8) \\
 balance\_scale   & 0.002 (7) & 0.007 (8) & 0.001 (4) & 0.001 (2)  & 0.001 (3) & 0.002 (6) & 0.002 (5)   & 0.001 (1) \\
 cars            & 0.073 (8) & 0.003 (4) & 0.010 (6) & 0.004 (5)  & \textbf{0.001} (1) & 0.002 (3) & 0.014 (7)   & 0.002 (2) \\
 mofn\_3\_7\_10     & 0.022 (8) & 0.003 (7) & 0.002 (5) & 0.001 (4)  & 0.001 (3) & 0.002 (6) & \textbf{0.001} (1)   & 0.001 (2) \\
 schizo          & 0.053 (8) & 0.004 (3) & 0.013 (6) & 0.008 (4)  & 0.004 (2) & \textbf{0.002} (1) & 0.012 (5)   & 0.017 (7) \\
 breast\_cancer   & 0.011 (5) & 0.051 (8) & 0.017 (6) & \textbf{0.003} (1)  & 0.017 (7) & 0.003 (2) & 0.007 (3)   & 0.007 (4) \\
 penguins        & 0.052 (8) & 0.006 (7) & 0.004 (6) & \textbf{0.000} (1)  & 0.000 (2) & 0.001 (3) & 0.002 (5)   & 0.002 (4) \\
 xd6             & 0.020 (8) & 0.002 (5) & 0.002 (7) & 0.001 (4)  & 0.001 (3) & 0.002 (6) & \textbf{0.000} (1)   & 0.000 (2) \\
 profb           & 0.077 (8) & 0.038 (7) & 0.006 (6) & 0.002 (2)  & 0.003 (4) & \textbf{0.001} (1) & 0.003 (5)   & 0.002 (3) \\
 monk2           & 0.004 (7) & 0.003 (4) & \textbf{0.001} (1) & 0.001 (3)  & 0.004 (6) & 0.001 (2) & 0.006 (8)   & 0.004 (5) \\
 saheart         & 0.079 (8) & 0.062 (7) & 0.011 (6) & 0.005 (5)  & \textbf{0.001} (1) & 0.003 (3) & 0.005 (4)   & 0.002 (2) \\
 tic\_tac\_toe     & 0.009 (8) & 0.007 (7) & 0.005 (6) & 0.002 (3)  & 0.002 (5) & 0.001 (2) & 0.002 (4)   & \textbf{0.001} (1) \\
 biomed          & 0.069 (8) & 0.019 (7) & 0.002 (3) & 0.001 (2)  & \textbf{0.001} (1) & 0.002 (4) & 0.011 (5)   & 0.012 (6) \\
 ionosphere      & 0.055 (8) & 0.008 (4) & 0.012 (7) & 0.010 (5)  & 0.004 (2) & \textbf{0.003} (1) & 0.004 (3)   & 0.011 (6) \\
 tokyo1          & 0.056 (8) & 0.017 (3) & 0.054 (7) & 0.018 (4)  & \textbf{0.001} (1) & 0.002 (2) & 0.020 (5)   & 0.021 (6) \\
 irish           & 0.070 (8) & 0.030 (7) & 0.011 (5) & 0.002 (4)  & 0.001 (2) & 0.002 (3) & 0.020 (6)   & \textbf{0.001} (1) \\
 glass2          & 0.105 (8) & 0.025 (6) & 0.035 (7) & 0.021 (5)  & \textbf{0.002} (1) & 0.006 (2) & 0.014 (3)   & 0.018 (4) \\
 bupa            & 0.065 (8) & 0.040 (7) & 0.027 (6) & 0.008 (5)  & \textbf{0.003} (1) & 0.003 (2) & 0.008 (4)   & 0.006 (3) \\
 parity5+5       & 0.017 (8) & 0.003 (7) & 0.002 (6) & 0.001 (5)  & 0.001 (3) & 0.001 (4) & \textbf{0.001} (1)   & 0.001 (2) \\
 germangss       & 0.086 (8) & 0.024 (6) & 0.012 (5) & 0.001 (2)  & \textbf{0.001} (1) & 0.003 (3) & 0.059 (7)   & 0.008 (4) \\
 prnn\_crabs      & 0.092 (8) & 0.011 (4) & 0.033 (6) & 0.042 (7)  & \textbf{0.001} (1) & 0.001 (2) & 0.026 (5)   & 0.010 (3) \\
\bottomrule
\end{tabular}
    \label{tab:mmd}
\end{table}

Figure \ref{fig:e2-priv} presents the results of the remaining metrics that focus on privacy and utility. Similar to the previous figure, the color of a bar indicates the average metric value. 

\textsc{Mia auc} metric quantifies the ability of a classifier to detect samples that were used to train a model. This metric reveals the risks related to the use of \textsc{smote} and Synthpop generators. Out of the other methods, \textsc{tvae} and \textsc{dimso-r} also show slightly higher scores, which, in the case of the \textsc{rae} loss function, may result from aiming to match the synthetic samples to the original ones. The use of \textsc{pca} with this loss function reduces the \textsc{mia} effectiveness. In the \textsc{dcr} metric, which is based directly on the sample's distances, \textsc{smote} is the only method that reveals the potential for private data leakage. Across the \textsc{dpi auc} results, all evaluated methods maintain values close to 50\%. The last metric evaluated the utility of synthetic data in classification tasks. A value of zero in the \textsc{mle auc diff} metric indicates that the classifier trained with synthetic data obtained the same \textsc{auc} as the classifier trained with original samples. The values close to zero are achieved by \textsc{smote}, Synthpop, and both \textsc{dimso} configurations. The use of synthetic data generated with \textsc{ctgan}, TabDiff, and Univariate generators usually leads to a drop in classification performance. This metric, similar to the \textsc{nmi} presented in Figure \ref{fig:e2-sim}, reveals that the simple Univariate baseline, despite obtaining high similarity in \textsc{ks} and \textsc{kld} metrics, does not reliably represent the original data distributions. The results of the \textsc{mle auc diff} metric are also presented in Table \ref{tab:mle}. Smote, Synthpop, and \textsc{dimso} are among the methods that frequently achieve the best scores in this metric.

\begin{table}[!htb]
    \centering
    \caption{The averaged results of \textsc{mle auc diff} metric describing the utility of synthesized data. Negative values indicate that the \textsc{auc} of a classifier trained with synthetic data is lower compared to a classifier trained with original data. The values in parenthesis indicate rank and the best result per dataset is emphasized with bold.}
    \vspace{0.5em}
    \setlength{\tabcolsep}{3pt}
\renewcommand{\arraystretch}{1.0}
\tiny
\begin{tabular}{l|rrrrrrrr}
\toprule
 dataset & CTGAN      & TVAE       & TabDiff    & Univariate & SMOTE      & Synthpop   & DiMSO-R & DiMSO-W    \\ \midrule
 allrep                   & -0.018 (7) & -0.003 (4) & -0.021 (8) & -0.015 (6) & \textbf{-0.001} (1) & -0.001 (2) & -0.002 (3)  & -0.007 (5) \\
 yeast                    & -0.165 (8) & -0.025 (3) & -0.152 (6) & -0.163 (7) & -0.014 (2) & -0.026 (4) & -0.042 (5)  & \textbf{-0.010} (1) \\
 hypothyroid              & -0.526 (8) & -0.109 (4) & -0.525 (7) & -0.496 (6) & -0.030 (2) & \textbf{-0.022} (1) & -0.156 (5)  & -0.103 (3) \\
 lawsuit                  & -0.199 (6) & -0.338 (7) & -0.198 (5) & -0.367 (8) & -0.006 (2) & -0.098 (4) & -0.020 (3)  & \textbf{-0.002} (1) \\
 backache                 & -0.075 (5) & -0.040 (4) & -0.076 (6) & -0.164 (8) & \textbf{0.032} (1)  & -0.080 (7) & 0.013 (2)   & -0.006 (3) \\
 balance\_scale            & -0.328 (8) & -0.065 (4) & -0.188 (6) & -0.294 (7) & -0.030 (3) & -0.090 (5) & \textbf{-0.004} (1)  & -0.021 (2) \\
 cars                     & -0.200 (7) & -0.111 (5) & -0.123 (6) & -0.268 (8) & \textbf{-0.035} (1) & -0.064 (3) & -0.084 (4)  & -0.059 (2) \\
 mofn\_3\_7\_10              & -0.298 (7) & -0.057 (4) & -0.160 (6) & -0.486 (8) & \textbf{-0.004} (1) & -0.084 (5) & -0.005 (2)  & -0.037 (3) \\
 schizo                   & -0.081 (8) & \textbf{0.036} (1)  & 0.030 (2)  & -0.053 (7) & -0.011 (5) & -0.005 (4) & 0.029 (3)   & -0.029 (6) \\
 breast\_cancer            & -0.106 (6) & 0.002 (4)  & -0.108 (7) & -0.174 (8) & -0.006 (5) & 0.006 (3)  & 0.022 (2)   & \textbf{0.038} (1)  \\
 penguins                 & -0.280 (7) & -0.011 (4) & -0.040 (6) & -0.435 (8) & \textbf{0.000} (1) & -0.016 (5) & -0.002 (3)  & -0.001 (2) \\
 xd6                      & -0.492 (7) & -0.076 (5) & -0.305 (6) & -0.523 (8) & \textbf{0.000} (1)  & -0.036 (3) & -0.000 (2)  & -0.052 (4) \\
 profb                    & -0.106 (6) & -0.079 (5) & -0.125 (7) & -0.166 (8) & -0.040 (4) & \textbf{-0.021} (1) & -0.034 (2)  & -0.038 (3) \\
 monk2                    & -0.489 (6) & -0.416 (5) & -0.529 (8) & -0.528 (7) & -0.222 (2) & -0.301 (4) & \textbf{-0.094} (1)  & -0.288 (3) \\
 saheart                  & -0.081 (6) & \textbf{0.029} (1)  & -0.120 (7) & -0.153 (8) & -0.004 (3) & -0.013 (5) & 0.019 (2)   & -0.010 (4) \\
 tic\_tac\_toe              & -0.531 (8) & -0.265 (4) & -0.410 (6) & -0.453 (7) & \textbf{-0.063} (1) & -0.197 (3) & -0.091 (2)  & -0.311 (5) \\
 biomed                   & -0.224 (7) & -0.050 (4) & -0.223 (6) & -0.403 (8) & \textbf{0.004} (1)  & -0.101 (5) & -0.035 (2)  & -0.047 (3) \\
 ionosphere               & -0.363 (7) & -0.041 (4) & -0.139 (6) & -0.542 (8) & -0.008 (2) & -0.038 (3) & \textbf{-0.004} (1)  & -0.087 (5) \\
 tokyo1                   & -0.487 (8) & -0.018 (3) & -0.029 (6) & -0.422 (7) & \textbf{-0.007} (1) & -0.009 (2) & -0.023 (4)  & -0.026 (5) \\
 irish                    & -0.263 (7) & -0.025 (5) & -0.127 (6) & -0.414 (8) & -0.001 (2) & \textbf{0.000} (1) & -0.007 (3)  & -0.015 (4) \\
 glass2                   & -0.394 (7) & -0.140 (3) & -0.432 (8) & -0.285 (6) & \textbf{-0.016} (1) & -0.148 (4) & -0.070 (2)  & -0.151 (5) \\
 bupa                     & 0.028 (5)  & \textbf{0.076} (1)  & -0.030 (8) & -0.007 (6) & 0.056 (2)  & -0.013 (7) & 0.054 (3)   & 0.030 (4)  \\
 parity5+5                & -0.501 (2) & -0.554 (7) & -0.532 (5) & -0.507 (3) & -0.646 (8) & -0.535 (6) & \textbf{-0.470} (1)  & -0.523 (4) \\
 germangss                & -0.201 (8) & -0.142 (6) & -0.092 (5) & -0.155 (7) & -0.015 (3) & \textbf{0.007} (1)  & -0.087 (4)  & -0.014 (2) \\
 prnn\_crabs               & -0.357 (6) & -0.248 (4) & -0.260 (5) & -0.466 (8) & -0.003 (2) & -0.097 (3) & -0.359 (7)  & \textbf{0.020} (1)  \\
\bottomrule
\end{tabular}
    \label{tab:mle}
\end{table}

\begin{figure}[!htb]
    \centering
    \includegraphics[width=0.49\linewidth,trim={0.6cm 0 0.6cm 0},clip]{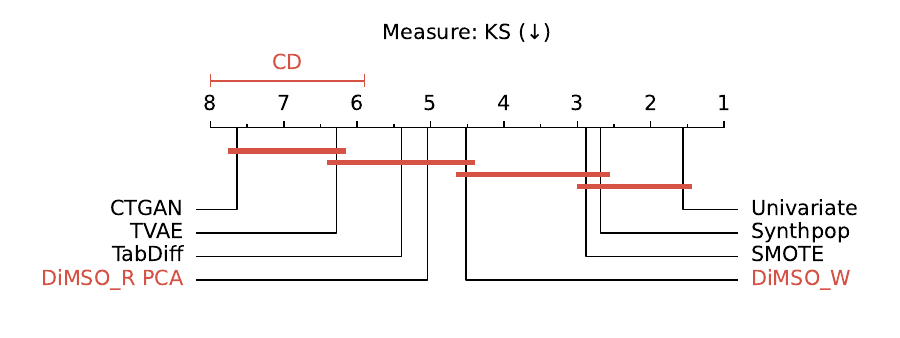}
    \includegraphics[width=0.49\linewidth,trim={0.6cm 0 0.6cm 0},clip]{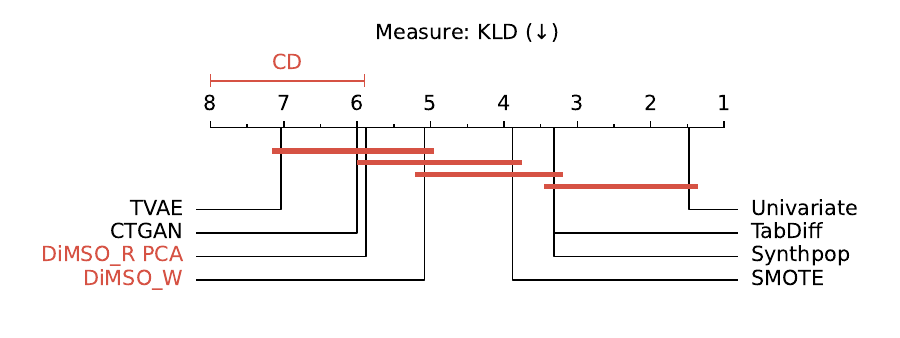}
    \includegraphics[width=0.49\linewidth,trim={0.6cm 0 0.6cm 0},clip]{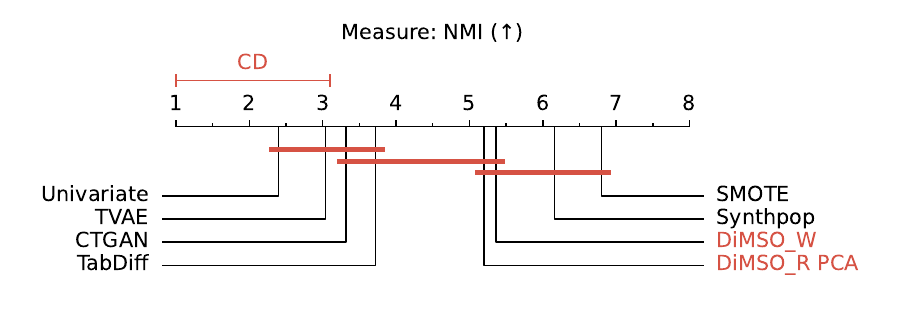}
    \includegraphics[width=0.49\linewidth,trim={0.6cm 0 0.6cm 0},clip]{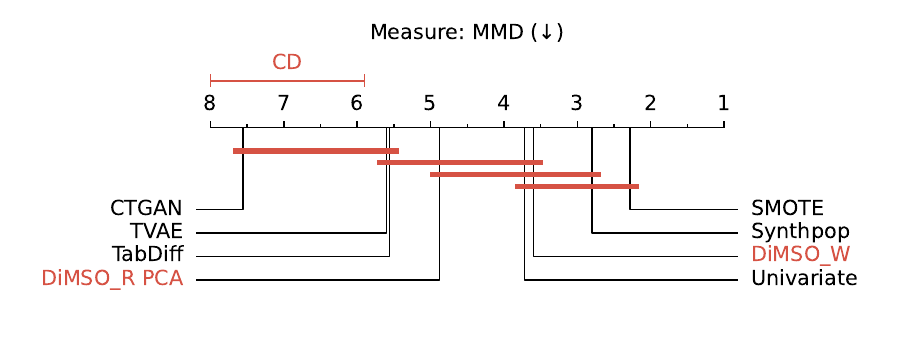}
    \caption{CD Diagrams of measures describing the distribution similarity. The results of methods connected with red lines are statistically related.}
    \label{fig:cd-sim}
\end{figure}

\begin{figure}[!htb]
    \centering
    \includegraphics[width=0.49\linewidth,trim={0.6cm 0 0.6cm 0},clip]{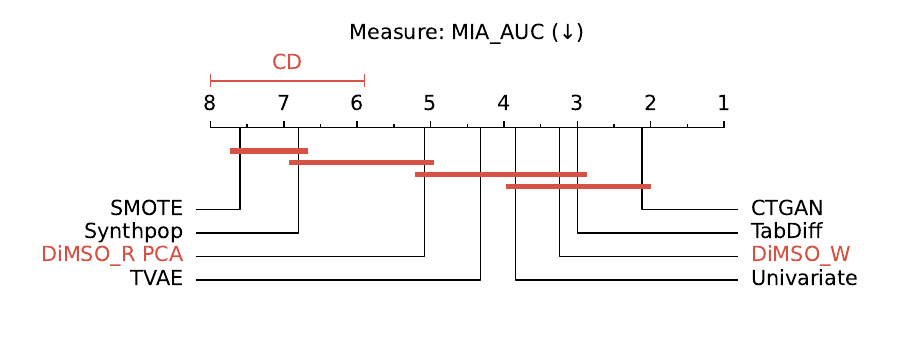}
    \includegraphics[width=0.49\linewidth,trim={0.6cm 0 0.6cm 0},clip]{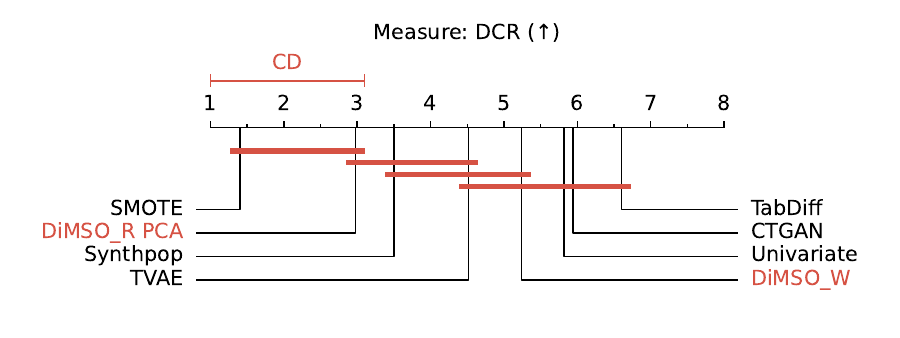}
    \includegraphics[width=0.49\linewidth,trim={0.6cm 0 0.6cm 0},clip]{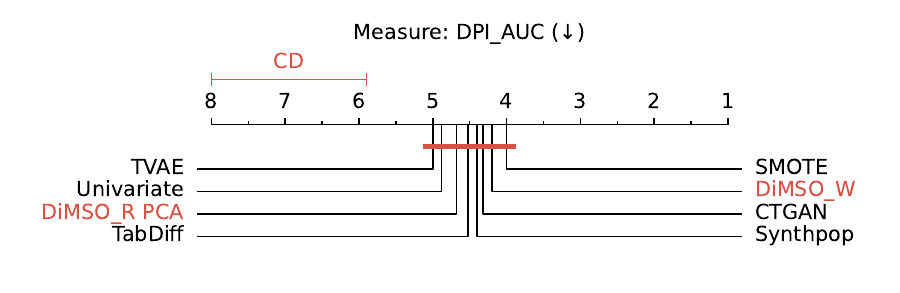}
    \includegraphics[width=0.49\linewidth,trim={0.6cm 0 0.6cm 0},clip]{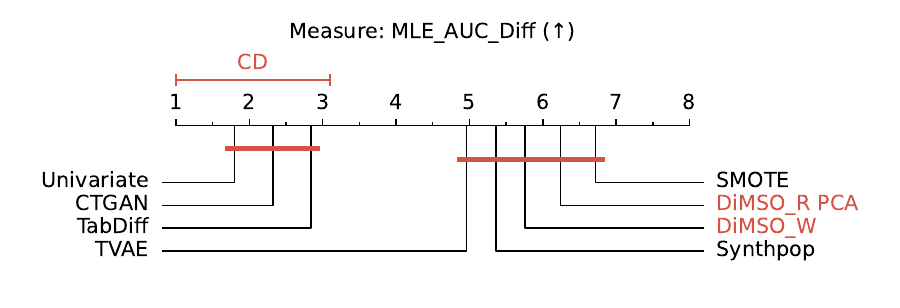}
    \caption{CD Diagrams of measures describing the privacy preservation and utility. The results of methods connected with red lines are statistically related.}
    \label{fig:cd-priv}
\end{figure}

The complete results were also subjected to statistical analysis with the non-parametric Friedman test with $\alpha = 0.05$. and the post-hoc Nemenyi test \cite{demvsar2006statistical}. The Critical Difference diagrams presenting the average ranks of the methods, as well as showing the statistical relationships of results, are presented in Figures \ref{fig:cd-sim} and \ref{fig:cd-priv}. The methods on the right side of an axis are obtaining the best ranks. The red line connecting the methods indicates that their results are statistically dependent. 

The CD diagrams were calculated for all metrics, however, it is important to remember that some of the measures are biased, e.g. \textsc{ks} and \textsc{kld} will score features independently and disregard the lack of feature-relationship preservation. Figure \ref{fig:cd-sim} shows that the proposed \textsc{dimso} is among the best performing methods in terms of \textsc{nmi} and \textsc{mmd} measures. In the case of \textsc{ks} and \textsc{kld}, \textsc{dimso-w} also ranks high and is related to Synthpop and \textsc{smote}. Generally, across similarity measures, simple methods, including Univariate baseline, often rank first.

Figure \ref{fig:cd-priv} shows that \textsc{dimso-w} is in the group of statistically related methods that rank best across all metrics. \textsc{Dpi auc} shows the lack of a significant difference between methods. In \textsc{mle auc diff}, there are two statistically related groups of methods, and \textsc{dimso}, with both configurations, is among in the better-ranked one.

The relationships of those metrics show that the assessment of generative methods is a complex task that can be approached from various directions. The results indicate that even though simple methods (\textsc{smote} and Synthpop) show good results in distribution similarity, they are surpassed by deep-learning approaches in terms of privacy preservation. The proposed \textsc{dimso} offers a valuable balance between the ability to model source distribution and provides mechanisms that protect potentially sensitive training data.

\subsection{The processing time comparison}

\begin{figure}[!htb]
    \centering
    \includegraphics[width=0.32\linewidth]{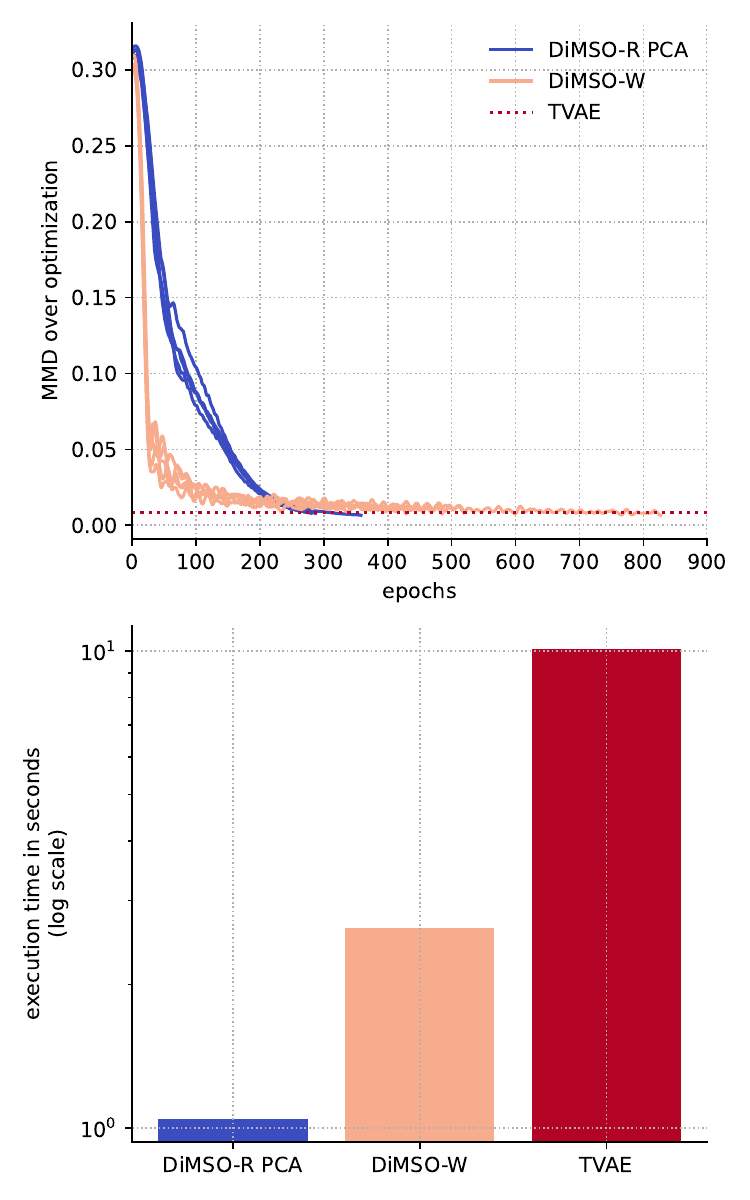}
    \includegraphics[width=0.32\linewidth]{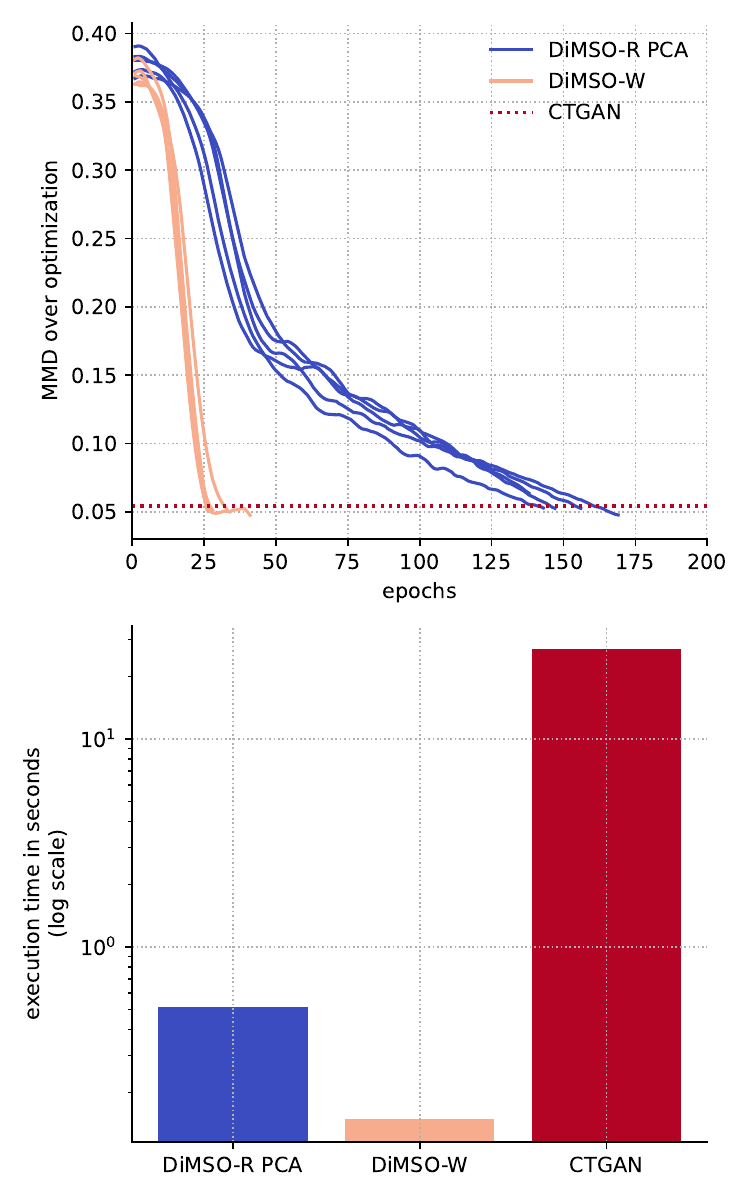}
    \includegraphics[width=0.32\linewidth]{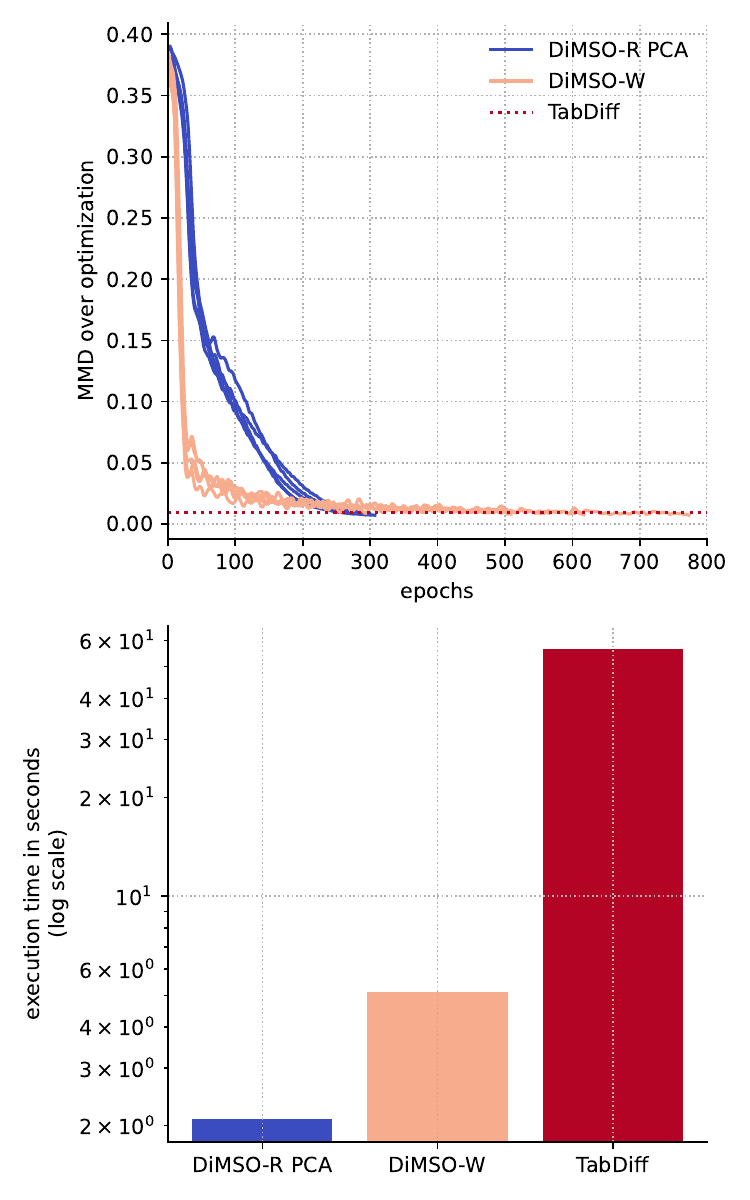}
    \caption{The values of \textsc{mmd} obtained with \textsc{dimso-r} \textsc{pca} (blue) and \textsc{dimso-w} (orange) across learning epochs (top rows). The reference \textsc{mmd} -- obtained with \textsc{tvae} (left), \textsc{ctgan} (center), TabDiff (right) -- is marked with dotted red line. Bottom row shows the average processing time of \textsc{dimso} and references on a logarithmic scale.}
    \label{fig:e4}
\end{figure}

The second experiment evaluated the processing time of \textsc{dimso} compared to reference deep learning solutions. The results of this experiment are presented in Figure \ref{fig:e4}. The experiment set the baseline \textsc{mmd} with a deep learning algorithm and monitored the distribution similarity while iteratively matching the distribution with \textsc{dimso}. The top plots show how the \textsc{mmd} of \textsc{dimso} evolved over training epochs in relation to deep learning solution (dotted red line), and the bottom section -- the processing time of reference (red) and \textsc{dimso} (blue and orange).

The learning curves show that \textsc{dimso-w} allows for faster reduction of \textsc{mmd} in the initial epochs. Since the target score for \textsc{ctgan} was set relatively high, this allowed \textsc{dimso-w} to converge in less than 50 epochs, resulting in a processing time of less than 0.15 seconds. \textsc{tvae} and TabDiff obtained scores that were more challenging to achieve with $\mathcal{L}_{WC}$. In those cases,  \textsc{dimso-r} converged quicker, in around 300 epochs for both references. The slower optimization in the case of the \textsc{rae} loss function in the initial steps results from the randomized selection of sample pairs, which makes the optimization face conflicting objectives. However, the simple pairwise distance reduction combined with \textsc{pca} to enhance privacy allowed \textsc{dimso-r} to achieve target \textsc{mmd} quicker than using Wasserstein-Covariance loss and in fewer epochs than the deep learning reference. The quick convergence of \textsc{dimso-w} in the case of \textsc{ctgan} resulted in its lowest processing time. In the case of \textsc{tvae} and TabDiff, \textsc{dimso-r} achieved target similarity quicker than \textsc{dimso-w}. In all cases, \textsc{dimso} took much less processing time compared to the deep learning solution. Table \ref{tab:e4-time} additionally presents the exact values of \textsc{mmd} that the methods achieved and the processing time of all methods.

\begin{table}[!htb]
    \centering
    \caption{The processing time of \textsc{dimso} in two configurations and reference deep learning approaches when optimizing the synthetic distribution towards an \textsc{mmd} set with a reference algorithm. The cell values indicate the average result, and the values in parentheses -- the standard deviation across replications.}
    \vspace{0.5em}
     \setlength{\tabcolsep}{5pt}
    \renewcommand{\arraystretch}{1.2}
    \scriptsize
    \begin{tabular}{c|c|lll}
    \toprule
     \textsc{reference} & \textsc{target similarity}  & \multicolumn{3}{c}{\textsc{optimization time [s]}}  \\ \midrule
     \multirow{2}{*}{TVAE} & \multirow{2}{*}{MMD = 0.0083 (0.0011)} & DiMSO-R PCA           & DiMSO-W           & TVAE           \\
      & & 1.047 (0.111) & 2.631 (0.728) & 10.096 (0.254)  \\ \midrule
     \multirow{2}{*}{CTGAN} & \multirow{2}{*}{MMD = 0.0542 (0.0056)} & DiMSO-R PCA          & DiMSO-W            & CTGAN          \\
      & & 0.515 (0.038) & 0.149 (0.028) & 26.971 (0.342) \\ \midrule
      \multirow{2}{*}{TabDiff} & \multirow{2}{*}{MMD = 0.0091 (0.0023)} & DiMSO-R PCA           & DiMSO-W         & TabDiff          \\
      & & 2.104 (0.235) & 5.097 (1.976) & 56.449 (1.160) \\ \bottomrule
    \hline
    \end{tabular}
    \label{tab:e4-time}
\end{table}

The Table, similar to the results presented in Figure \ref{fig:e4}, shows that \textsc{dimso} can achieve the target similarity between the synthetic and real-world distributions faster than the reference. Both approaches of \textsc{dimso} achieved the target \textsc{mmd} orders of magnitude faster than deep learning references, even though some replications required more epochs of \textsc{dimso} optimization. Such a difference results from the low computational cost of loss function calculation and the fast propagation through the fully connected network, making \textsc{dimso} ideal for systems with time and computational cost constraints.

\section{Conclusions and Limitations}

This work proposes a \textsc{dimso} tool for time-efficient synthetic data generation. The method uses a fully connected neural network and a randomized loss function combined with preprocessing designed for discrete tabular features to transform random high-dimensional Gaussian noise towards a target distribution. The mechanism of modeling the principal components employed in the method can further enhance data privacy protection. 

The presented research evaluated the proposed approach in two configurations across 25 real-world tabular datasets and compared it with traditional generative approaches and deep-learning solutions. The method achieved competitive results with state-of-the-art methods on metrics measuring distributional similarity between synthetic and original data, the privacy protection of original samples, and the utility in classification tasks. The time comparison experiment revealed that the proposed approach converges much faster than the deep learning references, which makes it a promising selection for time-constrained systems.

\begin{credits}
\subsubsection{\ackname} 
This work was supported by the statutory funds of the Department of Systems and Computer Networks, Faculty of Information and Communication Technology, Wrocław University of Science and Technology.

\subsubsection{\discintname}
The authors have no competing interests to declare that are
relevant to the content of this article.
\end{credits}

\bibliographystyle{splncs04}
\bibliography{bibliography}

\end{document}